# iBRF: Improved Balanced Random Forest Classifier

Asif Newaz, Md. Salman Mohosheu, MD. Abdullah al Noman
Islamic University of Technology (IUT)
Gazipur, Bangladesh
eee.asifnewaz@iut-dhaka.edu, salmanmohosheu@iut-dhaka.edu, alnoman6@iut-dhaka.edu

Dr. Taskeed Jabid
East West University
Dhaka, Bangladesh
taskeed@ewubd.edu

*Abstract*— Class imbalance poses a major challenge in different classification tasks, which is a frequently occurring scenario in many real-world applications. Data resampling is considered to be the standard approach to address this issue. The goal of the technique is to balance the class distribution by generating new samples or eliminating samples from the data. A wide variety of sampling techniques have been proposed over the years to tackle this challenging problem. Sampling techniques can also be incorporated into the ensemble learning framework to obtain more generalized prediction performance. Balanced Random Forest (BRF), RUSBoost, and SMOTE-Bagging are some of the popular ensemble approaches used in imbalanced learning. In this study, we propose a modification to the BRF classifier to enhance the prediction performance. In the original algorithm, the Random Undersampling (RUS) technique was utilized to balance the bootstrap samples. However, randomly eliminating too many samples from the data leads to significant data loss, resulting in a major decline in performance. We propose to alleviate the scenario by incorporating a novel hybrid sampling approach to balance the uneven class distribution in each bootstrap sub-sample. Our proposed hybrid sampling technique, when incorporated into the framework of the Random Forest classifier, termed as 'iBRF: improved Balanced Random Forest classifier', achieves better prediction performance than other sampling techniques used in imbalanced classification tasks. Experiments were carried out on 44 imbalanced datasets on which the original BRF classifier produced an average MCC score of 47.03% and an F1 score of 49.09%. Our proposed algorithm outperformed the approach by producing a far better MCC score of 53.04% and an F1 score of 55%. The results obtained signify the superiority of the iBRF algorithm and its potential to be an effective sampling technique in imbalanced learning.

## I. INTRODUCTION

Imbalanced data presents quite a hurdle in classification tasks as the learning algorithms get biased towards the majority class due to their accuracy-oriented design. They are trained to minimize the overall number of misclassifications, irrespective of the class. As a result, they perform well in predicting the majority class instances but they fail to correctly identify the instances belonging to the minority class as they are quite underrepresented. However, the minority class samples are usually more important and correctly predicting them is crucial in many applications such as medical diagnosis, fault diagnosis, or fraud detection. Most real-world datasets are more or less imbalanced. Especially in some critical applications, the data can be severely skewed. Therefore, necessary steps must be taken to address the class imbalance issue to develop a reliable prediction framework [1].

Resampling the data to balance the uneven class distribution prior to model training is an effective approach to deal with the imbalanced classification problem. Different resampling techniques have been proposed over the years which can be broadly classified into two categories: oversampling and undersampling [2]. Oversampling involves the generation of new minority-class samples, whereas, in undersampling, majority class instances are removed from the data. Random oversampling (ROS) and random undersampling (RUS) are two non-heuristic approaches in which the minority-class instances are duplicated to increase the number of samples and the majority-class instances are randomly eliminated to reduce the number of samples, respectively. Many other heuristic approaches have been developed to generate new synthetic samples rather than merely duplicating them [3]. SMOTE is one of the most popular oversampling techniques that use interpolation among nearby minority class instances to generate new samples [4]. The popularity and success of the technique have led to the development of a variety of modifications to the original algorithm [5]. Similarly, heuristic undersampling approaches have also been developed to strategically remove majority-class samples from the data, rather than randomly [6]. Hybridization between oversampling and undersampling approaches is also feasible and performs quite well [7]. These sampling techniques can be further incorporated into the ensemble learning framework to achieve better generalization. Ensemble algorithms are comprised of several weak learners, each trained on a different subset of the data. The predictions made by different weak learners, usually decision trees (DT), are aggregated to obtain a more robust classifier. These ensemble algorithms have reduced bias and variance and often provide better results than a single DT or other classifiers. The Balanced Random Forest (BRF) classifier is one such ensemble approach that uses RUS to resample each bootstrap subset of the data before training the DTs [8]. Other ensemble approaches use techniques such as ROS and SMOTE, producing algorithms like Over-Bagging, and SMOTE-Bagging, respectively. The sampling techniques can also utilized with the boosting framework. RUSBoost and Over-Boost are examples of such approaches.

Sampling techniques can effectively address the class imbalance issue and improve prediction performance. However, the success relies largely on how the data is resampled as well as other data intrinsic characteristics such as class overlapping, presence of noisy samples, imbalance ratio (IR), small disjuncts, etc. [9]. Techniques like ROS do not generate any new samples and thereby, do not add any new information for the model to learn. Although techniques like SMOTE and its variants alleviate this issue, the synthetic samples generated in the process do not always represent the minority class [10]. This occurs due to overlapping class regions and noisy samples around the decision boundary. Undersampling techniques like RUS indiscriminately remove majority-class samples from the data which can lead to the loss of valuable information. Especially in highly imbalanced scenarios, there is a major decline in performance even after using the sampling techniques. This is because when the data is highly skewed, too many samples need to be eliminated from the majority class or too many samples need to be generated from a small number of minority-class samples, which results in overfitting and loss of generalization. Hybridization between oversampling and undersampling can alleviate the situation to some extent as it creates a balance between the number of samples to be generated or eliminated. However, just merely merging techniques such as ROS and RUS does not solve the original issues associated with those approaches as mentioned earlier. Merging sampling techniques like RUS with the ensemble approaches can prove beneficial as they reduce the impact of such a huge loss of data by using a number of bootstrap subsets. However, each bootstrap subset when undersampled, still suffers from loss of information, resulting in the generation of poorly trained DTs. While an ensemble of them might outperform a single DT, the overall classification performance remains unsatisfactory.

To enhance the prediction performance, we propose a modification to the original BRF classifier in this study. We focus on the shortcomings of the sampling approaches and design a new hybrid framework that can overcome those limitations. Here, we first utilize the neighborhood cleaning (NC) rule algorithm to remove noisy majority-class samples from the data. Then we use the RUS approach to reduce the IR by randomly eliminating some samples from the majority class. Next, SMOTE is utilized to generate new synthetic samples to increase the number of minority class instances and balance the class distribution. These three approaches together form the hybrid sampling framework that we then integrate with the Random Forest (RF) architecture to benefit from the power of the ensemble and obtain better generalization. The RF classifier generates bootstrap subsets of the original data by randomly sampling with replacement. Each bootstrap subset is resampled using this hybrid sampling technique independently, allowing greater variation in the data. Individual DTs are then trained on these balanced bootstrap subsets and the predictions made by the models are aggregated, forming the final algorithm, termed 'iBRF'.

The boundary samples are usually the most difficult to identify as they obscure the regions between two classes. This also complicates the learning process. Since the minority-class samples are usually more important and rare, if the majority class samples in those overlapping regions are removed, it will ease the identification of those rare instances. The NC algorithm performs this exact task by clearing out these regions. This also reduces the IR and simplifies the generation process of new synthetic samples. SMOTE creates new samples in the region in between two nearby minority class samples. Since the NC algorithm has already eliminated noisy majority class samples from those regions, it becomes easier for the classifier to differentiate between the two classes. To avoid generating too many synthetic samples which can lead to overfitting, RUS is applied to randomly eliminate some samples from the majority class to further reduce the IR. A balance between the three approaches can provide a better sampling framework than using those techniques independently. Next, this hybrid framework is applied separately to each of the bootstrap subsets to balance the class distribution. As RUS arbitrarily removes samples from the data, some information might be lost in the bootstrap subsets. However, since we use an ensemble of 'n' (n=100 used in this study) different bootstrap subsets, some information may be lost in a few subsets but that information is present in other subsets. Thus the effect of information loss is minimized using the bagging process. Both SMOTE and NC similarly benefit from this bootstrapping step. This randomization introduces variations in the data and helps the model achieve better generalization. This proposed approach is tested on a wide range of imbalanced datasets and outperforms other sampling approaches by a large margin.

The rest of the article is structured as follows: Related works have been discussed in section- II. In section- III, we provide a detailed description of our proposed methodology. We discuss our experimental setup in section- IV. In section- V, we present the performance results and compare them with other sampling approaches. We conclude this article with section- VI by providing a summarization of the work and discussing some future research scopes

## II. LITERATURE REVIEW

Over the years, many different techniques have been introduced by researchers to address the class imbalance issue. This includes different variations of the SMOTE approach [5], the inclusion of evolutionary algorithms with the sampling techniques [11], a fusion between sampling and cost-sensitive learning [12], and hybridization among different sampling approaches [13]. In, [14] Gyorgy Kovac represented the performance of 85 SMOTE-variants on a wide range of imbalanced datasets. The author concluded that there was not a significant difference in performance among these numerous variations. While the popularity of the SMOTE approach led to the creation of many extensions of the original algorithm, the oversampling technique alone is usually not enough to obtain desirable performance [10]. The idea of hybridization seems more promising, and several such approaches have been developed. For instance, in [15], Xu et al. proposed a hybridization between M-SMOTE and ENN for medical imbalanced data and used the MCC score to compare the performance among different techniques. Other hybridization between different approaches is also possible and requires further research. One major issue with the sampling techniques is that they increase the variance, and the models suffer from loss of generalization [16]. The use of ensemble learning is a plausible solution in this regard and researchers have developed new approaches to incorporate ensemble methodologies in class imbalance problems. In [17], Dıez-Pastor et al. proposed an ensemble approach termed 'Random Balance' that randomly balances different subsets of the ensemble using the SMOTE algorithm to introduce more diversity. The authors later extended the algorithm to adapt to multiclass imbalanced scenarios [13]. In [18], Ribeiro et al. introduced a multi-objective optimization design to improve the performance of the ensemble approaches. They tested the performance of the model on a real-world anomaly detection problem and obtained better results. In another study 1 [19], Yang et al. proposed a hybrid ensemble classifier that combines density-based undersampling with cost-sensitive learning. The sampling techniques can be utilized with both bagging and boosting frameworks. However, just merely integrating the approaches with the ensemble techniques does not improve the prediction performance, as can be observed from the experimental results obtained from this study. The sampling technique first must be able to produce properly resampled data from which the separation between the two classes becomes easier. Only then the ensemble will be able to provide better generalization. In this regard, we propose a novel hybrid sampling approach to effectively balance the dataset before merging it with the ensemble framework. Our proposed approach has been evaluated on 44 different imbalanced scenarios and has demonstrated superior performance compared to other approaches, as evidenced by improvements in MCC, ROC-AUC, and F1 scores.

## III. METHODOLOGY

In this study, we present a novel ensemble sampling framework to effectively address the class imbalance problem and improve prediction performance. The outline of the proposed approach is illustrated in Fig. 1. The imbalanced data is first divided into 'n' bootstrap subsets. Since there is usually a limited number of minority-class samples available in the data, all these samples are taken in each bootstrap subset. Instances from the majority class are then randomly added with replacements to create the subsets. Each of these bootstrap subsets of the data is then balanced using the proposed hybrid sampling algorithm (illustrated in Fig. 2).

In the NC algorithm, for every sample in the training set, its k-nearest neighbors are located (k=3). Using these nearest neighbors, the original sample is classified. Now, if the original sample belongs to the majority class and is misclassified by its nearest neighbors– then that sample is removed from the training set. On the other hand, if the original sample belongs to the minority class and is misclassified by its nearest neighbors– then the nearest neighbors that belong to the majority class are removed from the training set. This way, the majority-class samples that can be considered noisy are identified and eliminated.

The iBRF algorithm was implemented using Scikit-learn and imbalanced-learn libraries. The implementation steps of the algorithm are as follows:

i. The data is first split into training and validation folds using a 5-fold stratified cross-validation strategy.

ii. 'n' number of bootstrap subsets is then generated from the training data including all minority class samples. The remaining portion is filled with samples from the majority class.

iii. The NC algorithm is applied to each bootstrap subset separately. The resultant resampled subsets have comparatively lower IR (different in different subsets) than the original data as some samples are removed in the process (depending on the data, usually around 10-20% of samples are eliminated).

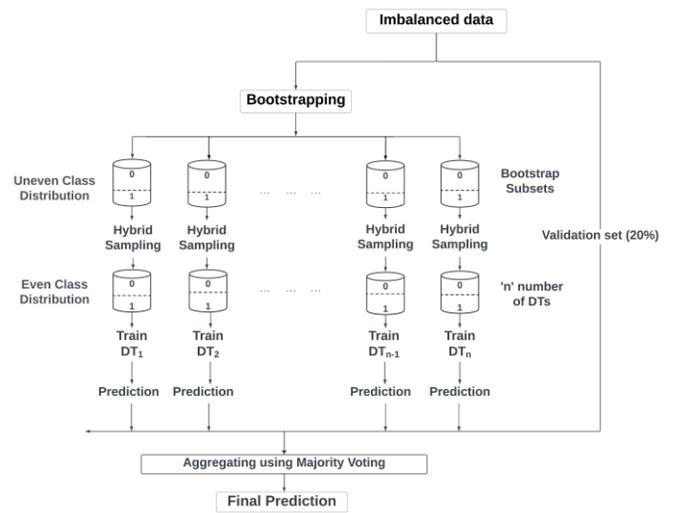

Fig. 1. iBRF: improved Balanced Random Forest classifier.

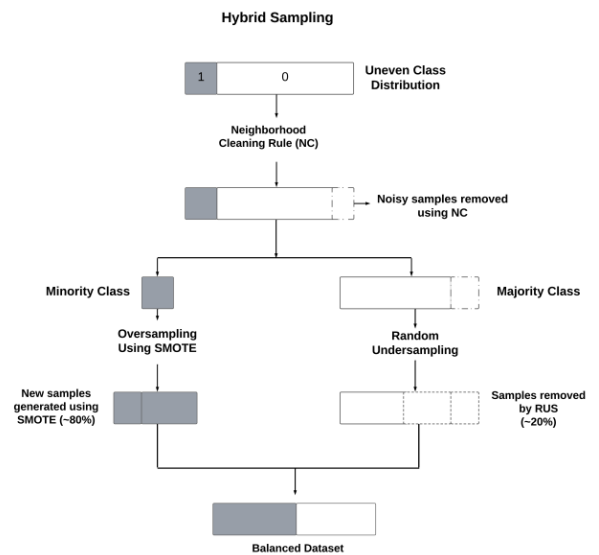

Fig. 2. The hybrid sampling process.

iv. The number of samples removed by the NC rule is usually limited. Consequently, a large number of minority class samples will need to be generated by the SMOTE approach to balance the class distribution. To avoid that, the RUS approach is applied to randomly eliminate around 20% majority-class samples from the data to lower the IR.

v. The SMOTE algorithm is then used to produce new minority class samples to balance each bootstrap subset.

vi. The percentage of samples to be generated or removed using the SMOTE and RUS approaches, respectively, can be considered as a hyperparameter and tuned for better results. In this study, however, we did not perform any hyperparameter tuning.

vii. Each bootstrap subset passes through steps iii, iv, and v, producing a balanced class distribution.

viii. Next, DTs are trained on these bootstrap subsets. The training and aggregation process follows the RF architecture of the scikit-learn library.

ix. The final prediction is made by the ensemble model on the validation set. The process is repeated for the other folds and the average of the prediction performance is calculated and reported here.

## IV. EXPERIMENTAL SETUP

To evaluate the performance of the proposed approach, it has been tested on 44 different imbalanced datasets. The datasets were collected from different sources including the KEEL and UCI repositories [20]. All the datasets are binary classification data with different degrees of imbalance. The performance of the iBRF classifier was compared with other popular techniques used in imbalanced learning. Eight different metrics were calculated during experimentation- MCC, ROC-AUC, F1-score, G-mean, sensitivity, specificity, precision, and accuracy. We mainly considered the MCC, ROC AUC, and F1-score for comparing the performance among different techniques. These are some of the most robust measures of classification performance. Some of the datasets on which the models are tested have a large class imbalance with only a handful of minority class samples available. Therefore, a 5-fold stratified cross-validation approach was undertaken for validation. Data resampling was performed only on the training folds to avoid any data leakage. The RF and SVM classifiers were utilized as the base learning algorithm for other sampling approaches. The default parameter values from the scikit-learn and imblearn libraries were utilized in obtaining the results.

## V. RESULTS AND DISCUSSION

In this section, the performance measures obtained from the iBRF algorithm have been presented. We compare our proposed approach with 14 popular sampling techniques used in imbalanced learning. The average performance obtained from the 44 imbalanced datasets is reported in this manuscript (Table 1). Performance metrics obtained from each dataset as well as dataset information are provided in a separate supplementary file due to space constraints.

Figure 3 presents a comparison of the MCC scores obtained from the iBRF algorithm with 7 other different sampling techniques which include 2 oversampling techniques (SMOTE and ADASYN), 3 undersampling techniques (RUS, NC, and CNN), and 2 hybrid sampling techniques (SMOTE-ENN and SMOTE-Tomek). As can be observed from the figure, our proposed approach outperforms all these techniques by a good margin. The SMOTE, RUS, and NC algorithms when used independently achieves an MCC score of 47.86%, 45.94%, and 46.09%, respectively. However, when these algorithms are merged together and integrated with the RF pipeline, the resultant approach performs significantly better producing an MCC score of 53.04%. This new approach also outperforms other hybrid sampling techniques.

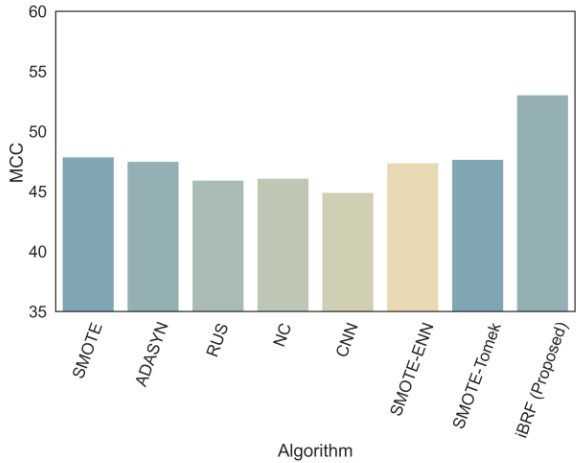

Fig. 3. Performance comparison of the proposed approach with other sampling techniques

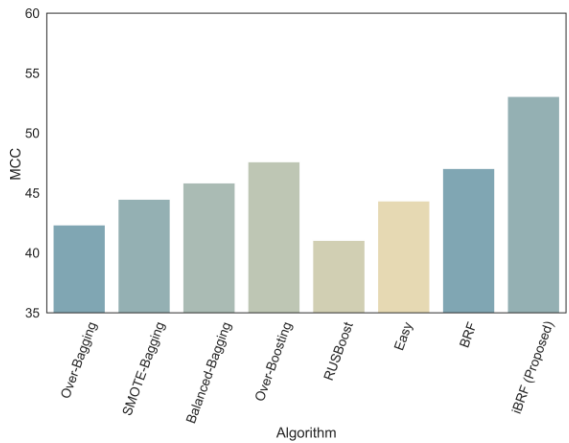

Fig. 4. Performance comparison of the proposed approach with other ensemble techniques

Figure 4 presents a comparison of the MCC scores obtained from the iBRF algorithm with 7 other ensemble sampling techniques used in imbalanced learning. This includes Over-Bagging, SMOTE-Bagging, Balanced-Bagging, Over Boosting, RUSBoost, Easy Ensemble, and BRF. Among these 7 algorithms, the over-Boost approach provided the highest MCC score of 47.58% and RUSBoost provided the lowest score of 41.04%. The BRF classifier produced a score of 47.03%. Compared to these techniques, our proposed iBRF classifier achieved an MCC score of 53.04%, which proves its superiority.

In terms of other metrics such as ROC-AUC, our model retains its top position as can be observed from Table 1. The algorithm attains a ROC-AUC score of 0.8226. Compared to this, the SMOTE approach attained a much lower ROC-AUC score of 0.74. The score obtained from the NC rule is even lower (0.72). Among the ensemble approaches, Over-Bagging

TABLE I

PERFORMANCE MEASURES OBTAINED FROM DIFFERENT APPROACHES ( IN PERCENTAGE )

| Methods | MCC | G-MEAN | ROC | Sensitivity | Specificity | Precision | Accuracy | F1-Score |
|---|---|---|---|---|---|---|---|---|
| NO SAMPLING | 44.047 | 51.476 | 69.416 | 41.688 | **97.144** | 58.864 | 92.891 | 45.247 |
| SMOTE | 47.869 | 64.136 | 74.225 | 54.580 | 93.870 | 53.198 | 91.276 | 51.656 |
| ADASYN | 47.496 | 63.394 | 73.758 | 53.948 | 93.567 | 51.841 | 90.948 | 50.390 |
| RUS | 45.937 | 78.625 | 80.514 | 81.843 | 79.185 | 41.580 | 79.456 | 48.569 |
| NC | 46.094 | 57.956 | 72.548 | 51.535 | 93.562 | **55.065** | 91.732 | 49.615 |
| CNN | 44.920 | 61.155 | 73.192 | 54.668 | 91.715 | 50.574 | 90.105 | 49.104 |
| SMOTE-ENN | 47.381 | 71.001 | 76.860 | 64.274 | 89.445 | 49.836 | 88.439 | 52.486 |
| SMOTE-TOMEK | 47.672 | 63.214 | 73.728 | 53.742 | 93.714 | 52.790 | 91.018 | 50.520 |
| OVER-BAGGING | 42.325 | 51.665 | 69.128 | 41.887 | 96.370 | 54.982 | **92.117** | 44.593 |
| SMOTE-BAGGING | 44.463 | 56.839 | 71.123 | 46.885 | 95.361 | 53.819 | 91.757 | 47.720 |
| BALANCED-BAGGING | 45.823 | 75.662 | 78.169 | 71.092 | 85.245 | 44.184 | 83.909 | 49.237 |
| OVER-BOOSTING | 47.590 | 69.911 | 76.193 | 62.021 | 90.366 | 49.660 | 88.606 | 51.650 |
| RUSBOOST | 41.037 | 68.597 | 74.313 | 62.245 | 86.382 | 43.113 | 84.640 | 46.212 |
| EASY ENSEMBLE | 44.325 | 78.674 | 80.425 | **82.604** | 78.246 | 39.554 | 78.687 | 47.105 |
| BRF | 47.031 | 79.244 | 81.044 | 81.776 | 80.312 | 42.065 | 80.514 | 49.095 |
| iBRF (proposed) | **53.042** | **79.923** | **82.260** | 78.931 | 85.589 | 48.709 | 85.880 | **55.002** |

provided the lowest score of 0.69, while Over-Boosting provided a score of 0.76. The BRF classifier achieved the highest ROC-AUC score of 0.81 among the ensemble approaches. However, our proposed modification to the algorithm improved its performance to 0.8226

In terms of the F1-score, the iBRF classifier achieved a score of 55%, which is a significant improvement over the BRF classifier (49%). Other ensemble approaches produced even poorer F1-score (47.72% from SMOTE-Bagging and 46.2% from RUSBoost). Some of the datasets utilized in this study have a large IR (more than 50). In such large imbalance cases, obtaining a high F1 score or MCC score is extremely difficult, even when the data has been resampled. However, compared to the other approaches, our proposed algorithm fared better, producing higher scores.

As can be observed from Table 1 and the discussion above, the proposed new algorithm is quite an improvement compared to the other sampling approaches. The hybridization among the 3 sampling techniques and the subsequent integration with the RF framework, allows the model to attain better generalization and achieve superior performance. Oversampling or undersampling by itself is not enough to shift the bias from the majority class in imbalanced scenarios. When only undersampling is performed on the data, the performance in the minority class (sensitivity) may improve significantly. However, this also results in a major decline in performance in the majority class (specificity). This is not desirable as many cases are incorrectly predicted. The number of misclassifications, irrespective of the class, is not captured by metrics such as G-mean. However, the MCC metric penalizes any misclassifications, and a higher value is an indication of better classification performance on all categories. Our proposed methodology achieved the highest MCC score as well as the ROC-AUC score compared to other state-of-the-art sampling techniques.

The hybrid sampling method establishes a balance between both oversampling and undersampling, while also ensuring that the samples that are generated are more representative of the minority class and majority of the samples that are eliminated are noisy, borderline samples. To obtain better generalization and reduce the effect of information loss, the process is merged with the bagging framework. Using boot strapping, a wide variety of subsets of the data is created, each with a slightly different data distribution. All these bootstrap subsets go through the hybrid sampling process, allowing variations in the resultant data. Each DT is then trained on a moderately different version of the data, resulting in variation in the learning process. The RF architecture inserts further randomization by allowing the trees to grow on a different subset of the features. Finally, the predictions made by individual DTs are aggregated, producing a more stable, robust, and generalized prediction model.

## VI. CONCLUSION

Sampling is a standard data preprocessing approach used to balance the class distribution when the data is skewed. In this study, we present a novel ensemble sampling approach to improve prediction performance. The model is built on top of the Random Forest architecture to benefit from the power of the ensemble. Weak learners by themselves are susceptible to bias and have high variance. By creating an ensemble of weak learners, better generalization can be achieved. However, the model remains susceptible to skewed class distribution and gets biased towards the majority class. Creating an ensemble does not solve the imbalance issue as each individual learner is still trained on imbalanced data. To alleviate the scenario, sampling techniques can be incorporated into the framework to balance each bootstrap sub-sample of the data. The weak learners will then be trained on a balanced subset of the data, resulting in a better prediction framework. The success of these approaches depends on how well the bootstrap subsets are resampled. The BRF classifier uses random undersampling to balance the subsets. Removing too many samples causes information loss, which greatly degrades the prediction performance. The BRF ensemble classifier mitigates this issue to some extent by using a large number of weak learners. However, the model still performs quite poorly when the IR is high which necessitates the removal of a large number of samples to balance the data. Models like SMOTE-Bagging or Over-Bagging, on the other hand, use oversampling techniques like SMOTE or ROS to generate new minority class samples to balance each bootstrap subset of the data. While the model might work well in low-imbalance scenarios, however, performance starts to drop in larger imbalances as too many samples need to be generated to balance the data, resulting in overfitting.

In our study, we propose a careful hybridization of three different sampling techniques to properly balance the class distribution. This hybrid sampling framework is then integrated into the RF architecture, producing the ensemble algorithm 'iBRF'. This proposed approach demonstrates a substantial improvement over other contemporary sampling techniques. This underscores the efficacy of the approach and its potential to serve as an effective sampling method in imbalanced classification tasks. In the future, we would like to extend our work and incorporate the hybrid sampling technique with the boosting framework. We would further test its performance on multiclass imbalanced scenarios and plan to explore other possible hybridizations to achieve better prediction performance.

Repository for supplementary files: https://github.com/newaz-aa/iBRF